\documentclass{article}

\usepackage[preprint]{neurips_2025}

\usepackage[utf8]{inputenc}
\usepackage[T1]{fontenc}
\usepackage{hyperref}
\usepackage{url}
\usepackage{booktabs}
\usepackage{amsmath}
\usepackage{amssymb}
\usepackage{graphicx}
\usepackage{subcaption}
\usepackage{float}
\usepackage{microtype}
\usepackage{xspace}
\usepackage{enumitem}

\setlist[itemize]{topsep=2pt, itemsep=1pt, parsep=0pt}

\newcommand{\tsa}{TSA\xspace}
\newcommand{\tlops}{TLOps\xspace}

\title{Adaptive Computation Depth via Learned Token Routing in Transformers}

\author{%
  Ahmed Abdelmuniem Abdalla Mohammed \\
  Independent Researcher \\
  \texttt{ahmed.abdelmuniem@gmail.com} \\
  \href{https://orcid.org/0009-0008-7410-6621}{ORCID: 0009-0008-7410-6621}
}
\begin{document}

\maketitle

% ============================================================
\begin{abstract}
Standard transformer architectures apply the same number of layers to every token
regardless of contextual difficulty. We present \textbf{Token-Selective Attention
(TSA)}, a learned per-token gate on residual updates between consecutive transformer
blocks. Each gate is a lightweight two-layer multi-layer perceptron (MLP) that produces a continuous halting
probability, making the mechanism end-to-end differentiable with 1.7\% parameter
overhead and no changes to the base architecture. Notably, \tsa learns
difficulty-proportional routing without any explicit depth pressure: even at
$\lambda{=}0$ (no depth regularisation), the task-loss gradient alone drives the
router to skip 20\% of token-layer operations. On character-level language modeling,
\tsa saved 14--23\% of token-layer operations (\tlops) across Tiny-Shakespeare and
enwik8 at $<$0.5\% quality loss. At matched efficiency, \tsa achieved 0.7\% lower
validation loss than early exit, and the learned routing transfers directly to
inference-time sparse execution for real wall-clock speedup.
\end{abstract}

\vspace{0.5em}

% ============================================================
\section{Introduction}

Transformer language models \citep{vaswani2017attention} apply a fixed number of
layers to every token in every sequence. This design trades per-token adaptability
for architectural simplicity. In practice, the trade-off is costly: a common token in
a predictable context requires far less processing than a rare token in a novel
construction, yet both receive identical compute at every layer of every forward pass.

The inefficiency is particularly consequential at inference scale. For large deployed
models, the dominant cost is the forward pass through all layers for all tokens. If a
significant fraction of tokens could exit early without quality loss, the savings
would translate directly to reduced latency and throughput gains.

Several approaches have addressed this problem. \citet{graves2016act} introduced
Adaptive Computation Time (ACT) for recurrent neural networks (RNNs), accumulating a halting probability across
recurrent steps. \citet{dehghani2019universal} extended the idea to depth-shared
transformer layers with the Universal Transformer. More recently,
\citet{raposo2024mixture} proposed Mixture-of-Depths (MoD), which routes tokens
through a fixed subset of layers using hard top-$k$ selection;
\citet{bae2025mor} introduced Mixture of Recursions, which applies recursive blocks
for a learned number of steps per token; and \citet{chen2025itt} presented the Inner
Thinking Transformer, which inserts additional computation steps at high-stakes
positions.

We present \textbf{Token-Selective Attention (TSA)}: a continuous soft gate on
residual updates, conditioned per token on its current hidden state. The mechanism is
architecturally minimal---a two-layer MLP per inter-block gap---and fully
differentiable, requiring no straight-through estimators, Gumbel sampling, or
reinforcement learning. Our contributions are:
\begin{itemize}
  \item A simple, differentiable token routing mechanism that gates residual
        updates softly per token per layer (\S\ref{sec:method}).
  \item Evidence that routing emerges from the task-loss gradient alone:
        at $\lambda{=}0$ (no depth regularisation), the router learns to
        skip 20\% of token-layer operations without any explicit depth
        pressure (\S\ref{sec:ablation_lambda}).
  \item Cross-dataset validation on character-level language modeling: 14--23\%
        token-layer operations saved across Tiny-Shakespeare and enwik8 at
        $<$0.5\% quality loss (\S\ref{sec:lm}, \S\ref{sec:enwik8}).
  \item Ablations showing robustness to $\lambda$ across two orders of
        magnitude (\S\ref{sec:ablation_lambda}), quality advantage over early
        exit at matched efficiency (\S\ref{sec:ablation_ee}), and real
        wall-clock speedup via sparse inference on commodity hardware
        (\S\ref{sec:ablation_wc}).
\end{itemize}

% ============================================================
\section{Method}
\label{sec:method}

\subsection{Architecture}

Let a pre-norm decoder-only transformer have blocks $f_0, f_1, \ldots, f_{L-1}$,
where each block applies multi-head self-attention and a feed-forward network (FFN) with
residual connections and LayerNorm \citep{ba2016layernorm}. In \tsa, a lightweight
router $r_l$ is inserted after each block $f_l$ for $l = 0, \ldots, L-2$.

Block $f_0$ is the \emph{stem} and always executes unconditionally: a bare token
embedding carries no contextual signal, making a routing decision at step zero
uninformative and potentially degenerate. The routing begins after the stem:

\begin{equation}
  h \leftarrow f_0(h), \quad
  p_l = r_l(h), \quad
  h \leftarrow f_{l+1}(h,\, p_l), \quad l = 0, \ldots, L-2.
\end{equation}

\begin{figure}[t]
  \centering
  \includegraphics[width=0.65\textwidth]{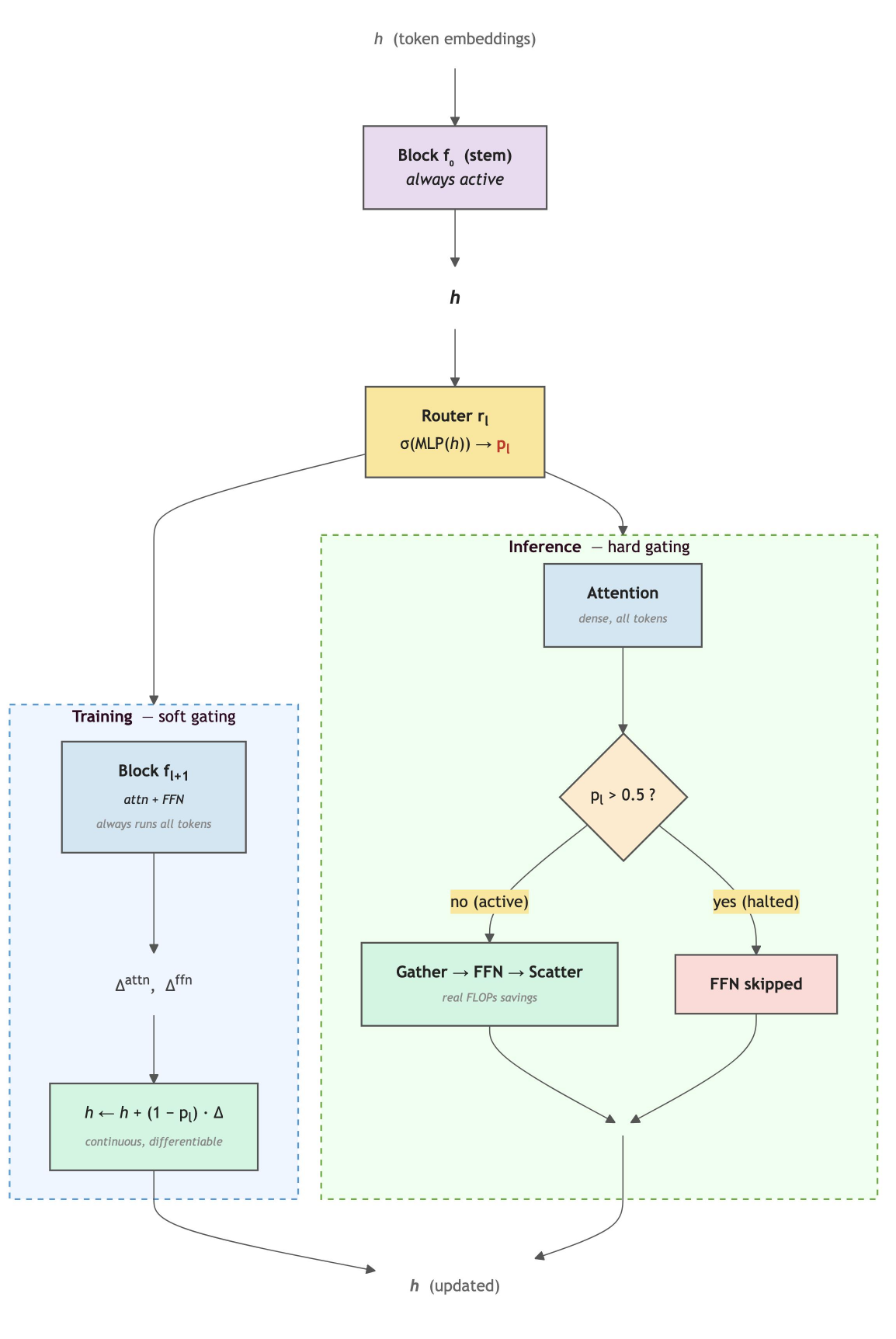}
  \caption{TSA dual-mode architecture. A router $r_l$ reads hidden state $h$
  and produces a per-token halting probability $p_l$. \textbf{Left (training):}
  all tokens always pass through attn + FFN; the residual update is soft-scaled
  by $(1{-}p_l)$, keeping the gate differentiable so the router learns.
  \textbf{Right (inference):} attention remains dense, but tokens with
  $p_l > 0.5$ skip the FFN entirely via gather/scatter, yielding real FLOPs
  savings. The stem block $f_0$ always executes unconditionally.}
  \label{fig:architecture}
\end{figure}

\noindent Figure~\ref{fig:architecture} illustrates the dual-mode mechanism:
soft gating during training (differentiable) and hard-threshold sparse execution
at inference (real FLOPs savings).

\subsection{Router Architecture}

Each router is a two-layer MLP with sigmoid output:
\begin{equation}
  r_l(h) = \sigma\!\bigl(W_l^{(2)}\,\mathrm{ReLU}(W_l^{(1)} h + b_l^{(1)}) + b_l^{(2)}\bigr),
  \quad r_l(h) \in (0,1)^{B \times T},
\end{equation}
where the hidden dimension is $d/4$ (floored at 16). Each router adds
$d^2/4 + d/2 + 1$ parameters; at $d=256$, $L=6$, this totals $\approx$83K on a 4.78M
parameter base model (1.7\% overhead).

The final bias $b_l^{(2)}$ is initialised to $-1.0$, giving
$\sigma(-1) \approx 0.27$ at initialisation. This bias prevents early collapse to
``halt everything'' before the model has learned useful representations.

\subsection{Gated Block Update}

For each routing decision $l = 0, \ldots, L{-}2$, the gated update of block
$f_{l+1}$ is:
\begin{align}
  h &\leftarrow h + (1 - p_l) \odot \Delta_{l+1}^{\mathrm{attn}}(h), \\
  h &\leftarrow h + (1 - p_l) \odot \Delta_{l+1}^{\mathrm{ffn}}(h),
\end{align}
where $p_l \in (0,1)^{B \times T}$ is broadcast over the model dimension $d$, and
$\Delta_{l+1}^{\mathrm{attn}}$, $\Delta_{l+1}^{\mathrm{ffn}}$ are the pre-norm
attention and feed-forward residual deltas of block $f_{l+1}$ respectively. When $p_l = 0$, the update is identical
to the standard transformer. When $p_l = 1$, the state is unchanged---the block is
skipped. The interpolation is smooth, preserving gradient flow through $p_l$ during
training.

\subsection{Depth Regularisation}

Without any incentive to halt, routers default to $p_l \approx 0$ and \tsa reduces
to a standard transformer with extra parameters. We added a depth regularisation term
that gently encourages early halting:
\begin{equation}
  \mathcal{L}_\mathrm{depth} = \lambda \cdot \frac{1}{L-1}
  \sum_{l=0}^{L-2} \overline{1 - p_l},
\end{equation}
where $\overline{1-p_l}$ is the mean active fraction at layer $l$ (averaged over
batch and sequence position). The total training loss is
$\mathcal{L} = \mathcal{L}_\mathrm{task} + \mathcal{L}_\mathrm{depth}$. We used
$\lambda = 0.001$ for language experiments; Section~\ref{sec:ablation_lambda}
demonstrates that \tsa is robust across $\lambda \in [0,\, 0.1]$.

\subsection{Compute Metric}

We measure compute using \textbf{token-layer operations (\tlops)}: for each block,
\tlops equals the number of tokens processed at that block. The mean active fraction
across routing decisions is:
\begin{equation}
  \alpha = \frac{1}{L-1} \sum_{l=0}^{L-2} \overline{1-p_l}.
\end{equation}
\tlops savings relative to the fixed-depth baseline are:
\begin{equation}
  \Delta = 1 - \frac{1 + (L-1)\,\alpha}{L}.
\end{equation}
The stem block (always active) is included in both numerator and denominator,
making $\Delta$ a conservative estimate.

\emph{Note on training compute.} During training, all layers execute fully: the gate
scales residual updates but does not skip computation. \tlops therefore measures the
effective contribution of each layer to the final representation, not actual FLOPs
saved. At inference, sparse-TSA (Section~\ref{sec:ablation_wc}) exploits
low-contribution positions via gather/scatter to achieve real compute savings.

% ============================================================
\section{Experiments}

\subsection{Synthetic Algorithmic Tasks}
\label{sec:synth}

\paragraph{Setup.}
We used decoder-only transformers trained on \textbf{copy} and \textbf{sort} tasks
over length-10 sequences from a 32-token vocabulary. Inputs followed the format
\texttt{[BOS] src [SEP] tgt [EOS]} with loss masked on source tokens. Both baseline
and \tsa used $d{=}128$, $L{=}6$, $H{=}4$, $d_\mathrm{ff}{=}512$ (baseline:
1.20M params; \tsa: 1.22M params, $+$1.7\%). Training employed AdamW
\citep{loshchilov2019adamw} with $\beta{=}(0.9,0.95)$, $\mathrm{lr}{=}3{\times}10^{-4}$,
$\lambda_\mathrm{wd}{=}0.1$ for 10K gradient steps on 10K training sequences.
We report token-level sequence accuracy on 1K held-out sequences.

\paragraph{Results.}

\begin{table}[h]
\centering
\caption{Synthetic Task Results ($d{=}128$, $L{=}6$, Toy Vocabulary)}
\label{tab:synthetic}
\small
\begin{tabular}{llcccc}
\toprule
Task & Model & Accuracy & Perplexity & Active Frac $\alpha$ & TLOps saved$^\dagger$ \\
\midrule
Copy & Baseline & 1.0000 & 1.012 & 1.000 & --- \\
Copy & \textbf{TSA}  & \textbf{1.0000} & \textbf{1.012} & \textbf{0.341} & \textbf{54.9\%} \\
\midrule
Sort & Baseline & 0.9915 & 1.079 & 1.000 & --- \\
Sort & \textbf{TSA}  & 0.9878 & 1.087 & \textbf{0.730} & \textbf{22.5\%} \\
\bottomrule
\end{tabular}
\\[3pt]
{\footnotesize $^\dagger$ TLOps saved $= 1 - (1 + (L-1)\,\alpha)/L$, including the mandatory stem block.}
\end{table}

The routing pattern directly reflected task difficulty. Copy is an identity mapping:
the router learned that nearly all tokens were fully determined after the stem block
($\alpha = 0.341$; 54.9\% overall TLOps saved).
Sort requires comparison and permutation, yielding $\alpha = 0.730$---more compute
where the task genuinely demanded it. This difficulty-proportional allocation emerged
without any explicit supervision about task identity or difficulty.

\subsection{Character-Level Language Modeling}
\label{sec:lm}

\paragraph{Setup.}
We trained on Tiny-Shakespeare \citep{karpathy2015charray} (1.1M characters, 65-char
vocabulary, 80/10/10 train/val/test split). Both models used $d{=}256$, $L{=}6$,
$H{=}8$, $d_\mathrm{ff}{=}1024$, context length 128. Training employed AdamW with
cosine learning rate schedule, batch size 64, for 5{,}000 gradient steps (baseline:
4.78M params; \tsa: 4.86M params, $+$1.7\%). Token embeddings were initialised
without a padding index: character index 0 is the newline character
(\textasciitilde{}8\% of the corpus), whose embedding gradient must not be zeroed.

\paragraph{Results.}

\begin{table}[h]
\centering
\caption{Language Modeling Results ($d{=}256$, $L{=}6$, Tiny-Shakespeare)}
\label{tab:language}
\small
\begin{tabular}{lcccccc}
\toprule
Model & Params & Val Loss & BPC & $\alpha$ & TLOps saved \\
\midrule
Baseline & 4,782,336 & \textbf{1.4422} & \textbf{2.0807} & 1.000 & --- \\
\textbf{TSA}      & 4,864,901 & 1.4482        & 2.0893        & \textbf{0.726} & \textbf{22.8\%} \\
\bottomrule
\end{tabular}
\\[3pt]
{\footnotesize Val loss increase: $+$0.006 nats ($+$0.4\% relative). BPC = bits-per-character.}
\end{table}

\begin{figure}[t]
  \centering
  \begin{subfigure}[b]{0.48\textwidth}
    \includegraphics[width=\textwidth]{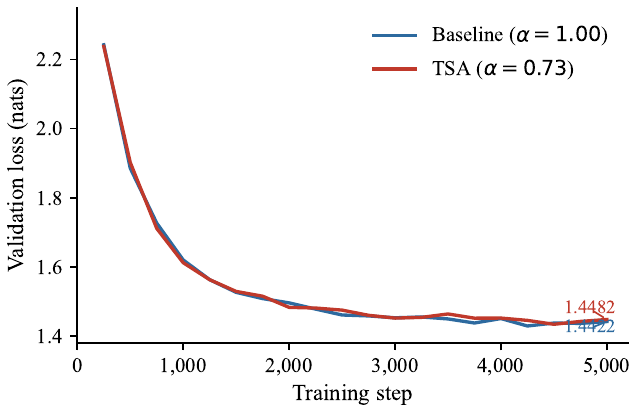}
    \caption{Validation loss vs.\ training step.}
    \label{fig:loss_steps}
  \end{subfigure}
  \hfill
  \begin{subfigure}[b]{0.48\textwidth}
    \includegraphics[width=\textwidth]{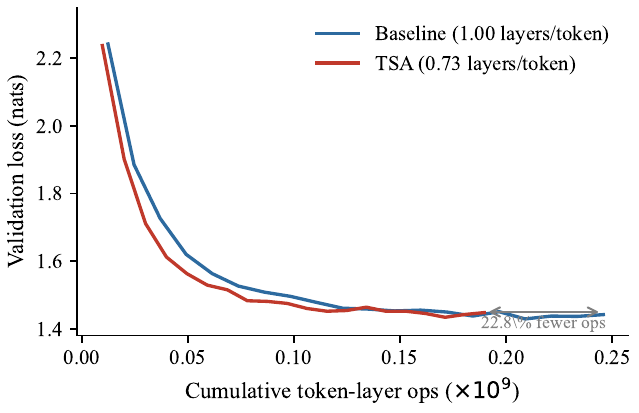}
    \caption{Validation loss vs.\ cumulative \tlops ($\times 10^9$).}
    \label{fig:loss_compute}
  \end{subfigure}
  \caption{TSA (red) and Baseline (blue) on Tiny-Shakespeare. Left: equivalent
  convergence speed. Right: TSA reaches the same loss for 22.8\% fewer
  token-layer operations.}
  \label{fig:language}
\end{figure}

\tsa achieved $\alpha = 0.726$: 22.8\% of token-layer operations saved at a cost
of 0.006~nats ($+$0.4\%) in validation loss.
Both models reached all convergence thresholds at identical step counts, indicating
\tsa did not impede convergence (Figure~\ref{fig:language}).
The \tsa curve lies consistently to the left on the compute axis, confirming that
savings remained stable throughout training.

% ============================================================
\subsection{Cross-Dataset Validation: enwik8}
\label{sec:enwik8}

To test whether \tsa generalises beyond a single corpus, we trained on
enwik8 \citep{hutter2006prize}: the first $10^8$~bytes of English Wikipedia
(raw XML, 6{,}064 unique characters). This corpus is substantially more diverse
than Shakespeare---it contains markup, multilingual text, tables, and mathematical
notation. We used $d{=}256$, $L{=}6$, $H{=}8$, $d_\mathrm{ff}{=}1024$, context
length 256, batch size 64, for 5{,}000 steps (6.35M params baseline; 6.43M \tsa).
Experiments were conducted on Apple M1~Pro using
MLX \citep{mlx2023}.

\begin{table}[h]
\centering
\caption{enwik8 Results ($d{=}256$, $L{=}6$, Context 256)}
\label{tab:enwik8}
\small
\begin{tabular}{lccccc}
\toprule
Model & Val Loss & BPC & $\alpha$ & TLOps saved & ms/step \\
\midrule
Baseline & 1.2826 & 1.8504 & 1.000 & --- & 585 \\
\textbf{TSA} & \textbf{1.2774} & \textbf{1.8429} & \textbf{0.833} & \textbf{13.9\%} & 599 \\
\bottomrule
\end{tabular}
\\[3pt]
{\footnotesize TSA quality vs.\ Baseline: $-$0.4\% (TSA is marginally better; within noise).}
\end{table}

\tsa achieved $\alpha = 0.833$ on enwik8, more conservative than Shakespeare's
$\alpha = 0.726$. The router allocated more compute on the structurally diverse
Wikipedia corpus while still saving 13.9\% of \tlops at no quality cost.
Both conditions reached all convergence thresholds ($\leq$2.5, $\leq$2.0, $\leq$1.8 BPC)
at identical steps (500, 750, 1{,}000). The cross-dataset result confirms that the
routing mechanism learned a content-dependent signal rather than overfitting to
corpus-specific patterns. Training curves are presented in
Figure~\ref{fig:enwik8_curves} (Appendix).

% ============================================================
\subsection{Ablation: Depth Regularisation Sensitivity}
\label{sec:ablation_lambda}

\begin{figure}[t]
  \centering
  \begin{subfigure}[b]{0.48\textwidth}
    \includegraphics[width=\textwidth]{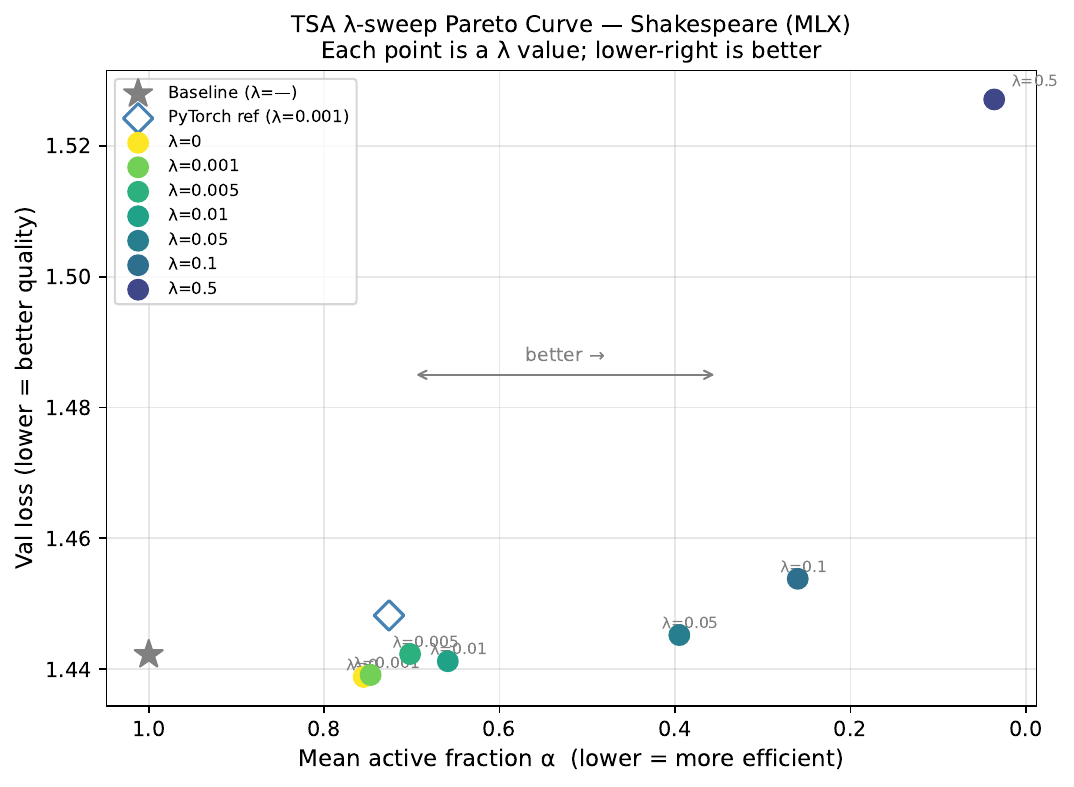}
    \caption{$\lambda$ sweep: val loss vs.\ active fraction.}
    \label{fig:ablation_lambda}
  \end{subfigure}
  \hfill
  \begin{subfigure}[b]{0.48\textwidth}
    \includegraphics[width=\textwidth]{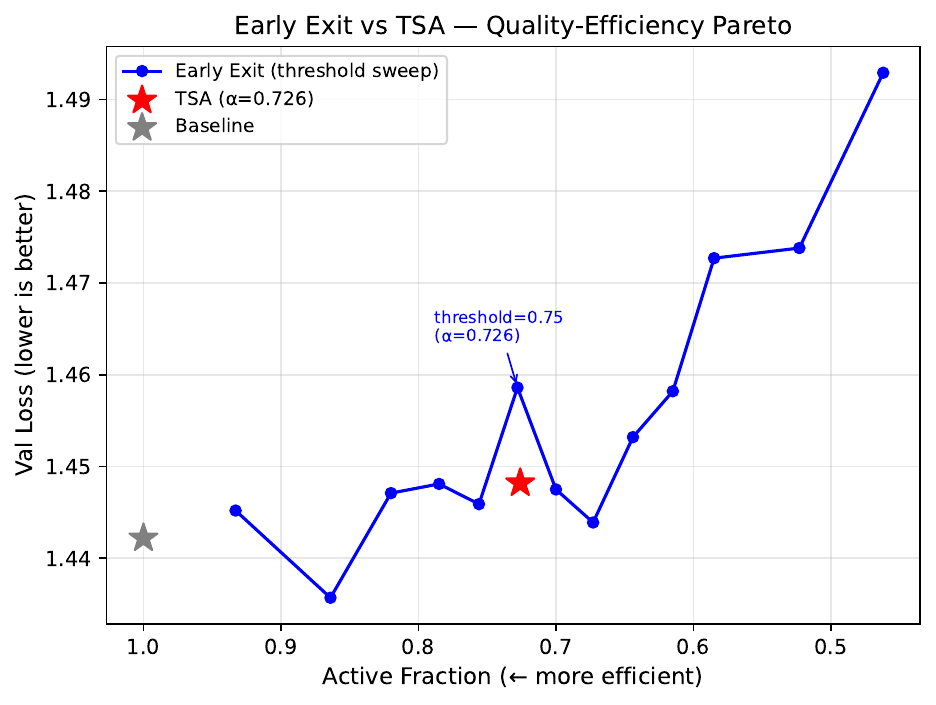}
    \caption{Early exit vs.\ \tsa Pareto curves.}
    \label{fig:ablation_ee}
  \end{subfigure}
  \caption{Ablation studies on Tiny-Shakespeare. \textbf{Left:} \tsa is robust
  across $\lambda \in [0,\, 0.1]$; quality range is 0.015~nats (1.04\%). Even
  $\lambda{=}0$ produces meaningful routing. \textbf{Right:} \tsa (red star)
  dominates the early exit threshold sweep (blue) at matched $\alpha \approx 0.726$.}
  \label{fig:ablations}
\end{figure}

We swept $\lambda \in \{0,\, 0.001,\, 0.005,\, 0.01,\, 0.05,\, 0.1,\, 0.5\}$ on
Tiny-Shakespeare with all other hyperparameters fixed at the values in
\S\ref{sec:lm}. Figure~\ref{fig:ablation_lambda} shows the
quality-efficiency Pareto curve; full results are presented in
Table~\ref{tab:lambda_full} (Appendix).

Three findings emerged.
\emph{(i)~$\lambda{=}0$ still routes:} without any explicit depth pressure, the
router learned to save 20.4\% of \tlops ($\alpha = 0.755$) via the task-loss gradient
alone. This is the central finding: the gating multiplication
$h \mathrel{+}= (1{-}p_l) \odot \Delta$ provides an intrinsic learning signal---when
a layer's residual update is noisy or redundant, the gradient favours increasing $p_l$
to attenuate the update, even without regularisation. The router thus acts as a
learned noise gate.
\emph{(ii)~Robustness:} across $\lambda \in [0,\, 0.1]$, the quality range was only
0.015~nats (1.04\% relative)---\tsa does not require precise tuning of $\lambda$.
\emph{(iii)~$\lambda{=}0.05$ is Pareto-optimal:} 50.4\% \tlops saved at $<$0.5\%
quality loss, 2.4 times as efficient as the default $\lambda{=}0.001$ with negligible
quality cost.
The stability boundary is $\lambda < 0.5$; at $\lambda{=}0.5$ the active fraction
collapsed to 0.036 and quality degraded by 5.9\%.

% ============================================================
\subsection{Ablation: Comparison With Early Exit}
\label{sec:ablation_ee}

The early exit approach \citep{elbayad2020depth} is the canonical inference-time
baseline for adaptive-depth transformers. We trained an early exit model with
$N$~auxiliary exit classifiers (one per block, tied embedding output heads, uniform
mean cross-entropy loss across all exits) on the same Shakespeare setup. At inference,
a token exited when its maximum softmax probability exceeded a confidence threshold.
We swept 13 thresholds and selected the one yielding $\alpha \approx 0.726$ to match
the \tsa operating point. Full threshold data are presented in
Table~\ref{tab:ee_full} (Appendix).

\begin{table}[h]
\centering
\caption{Comparison at Matched Active Fraction ($\alpha \approx 0.726$, Shakespeare)}
\label{tab:early_exit}
\small
\begin{tabular}{lccccc}
\toprule
Model & Val Loss & BPC & $\alpha$ & TLOps saved \\
\midrule
Baseline       & 1.4422 & 2.0807 & 1.000 & ---    \\
\textbf{TSA}   & 1.4482 & 2.0893 & 0.726 & 22.8\% \\
Early Exit     & 1.4586 & 2.1043 & 0.728 & 22.7\% \\
\bottomrule
\end{tabular}
\end{table}

\tsa's router operates at both train and inference time, learning routing decisions
end-to-end via the task-loss gradient. Early exit trains identically to the baseline
and applies routing only at inference via a separate confidence threshold. At matched
active fraction, \tsa achieved 0.71\% lower validation loss than early exit
(Table~\ref{tab:early_exit}), suggesting that end-to-end learned routing produces
better routing decisions than post-hoc confidence thresholding. At more conservative
thresholds (higher $\alpha$, fewer tokens skipped), early exit quality improves---as
expected, since fewer routing decisions are made---but the comparison at matched
$\alpha$ isolates routing quality from routing aggressiveness
(Table~\ref{tab:ee_full}).

% ============================================================
\subsection{Wall-Clock Throughput}
\label{sec:ablation_wc}

Soft gating (\S\ref{sec:method}) multiplies all residuals by $(1{-}p_l)$ but does
not skip any computation. To translate \tlops savings to wall-clock speedup, we
implemented \textbf{sparse-TSA}: at inference, tokens with $p_l > 0.5$ were excluded
from the FFN via gather/scatter operations; attention computation remained dense
to preserve exact key-value (KV) semantics, though the attention residual update was gated by
the same binary mask as the FFN. We benchmarked on Apple M1~Pro using
MLX \citep{mlx2023} with batch=64, seq=256, 30~warmup $+$ 200~timed forward
passes per configuration. Full data are presented in Table~\ref{tab:wc_full}
(Appendix).

\begin{table}[h]
\centering
\caption{Wall-Clock Throughput (M1 Pro, MLX, Batch=64, Seq=256)}
\label{tab:wallclock}
\small
\begin{tabular}{lcccc}
\toprule
Condition & $\alpha$ & TLOps saved & ms/step & Speedup \\
\midrule
Baseline        & ---   & ---    & 113.2 & $1.000\times$ \\
Soft-TSA        & any   & ---    & $\sim$114.4 & $\sim$$0.99\times$ \\
Sparse-TSA      & 0.726 & 22.8\% & 110.7 & $\mathbf{1.023\times}$ \\
Sparse-TSA      & 0.833 & 13.9\% & 113.1 & $1.000\times$ \\
\bottomrule
\end{tabular}
\\[3pt]
{\footnotesize Soft-TSA overhead ($\sim$1\%) is flat across all $\alpha$---the router
is negligible cost. Sparse-TSA speedup requires batch $\geq 64$.}
\end{table}

Soft gating added $\sim$1\% overhead regardless of~$\alpha$---the router
represented negligible cost. Sparse-TSA was faster than the baseline for
$\alpha \leq 0.83$: 2.3\% speedup at $\alpha{=}0.726$ (Shakespeare), break-even at
$\alpha{=}0.833$ (enwik8). The speedup required batch $\geq 64$; at batch$=$1,
CPU--GPU synchronisation dominated.

% ============================================================
\section{Analysis and Limitations}

\paragraph{What did the router learn?}
On synthetic tasks, routing was interpretable: copy halted maximally (identity
requires no deep computation); sort halted moderately (comparison needs depth).
On language, the router was more conservative on enwik8 ($\alpha = 0.833$)
than Shakespeare ($\alpha = 0.726$), consistent with Wikipedia's greater
structural diversity.

\begin{figure}[t]
  \centering
  \includegraphics[width=\textwidth]{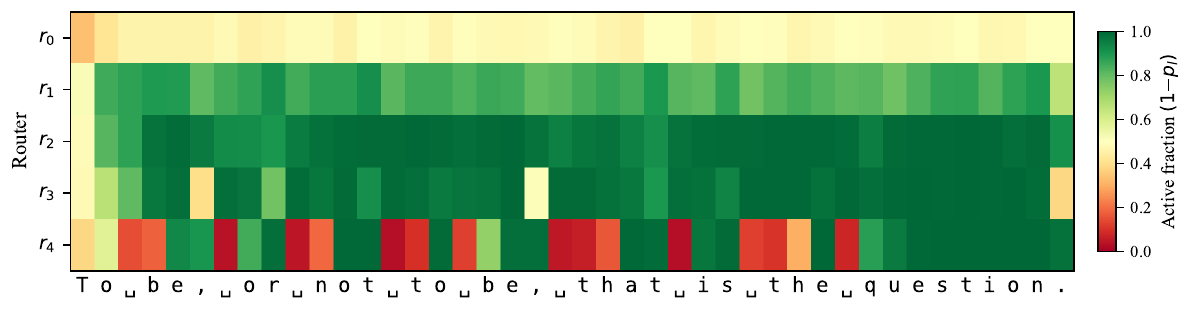}
  \caption{Per-token active fraction $(1{-}p_l)$ across five routing decisions
  on a Shakespeare passage. Green: fully active; red: nearly halted.
  Early routers stay permissive; later routers selectively gate predictable
  tokens (spaces, punctuation, common characters) while preserving
  computation for content-bearing positions.}
  \label{fig:routing_heatmap}
\end{figure}

Figure~\ref{fig:routing_heatmap} shows per-token routing decisions across a
sample Shakespeare passage. Early routers ($r_0$, $r_1$) keep most tokens
active, while later routers ($r_3$, $r_4$) exhibit selective gating:
punctuation, spaces, and predictable characters are attenuated more
aggressively than content-bearing characters in mid-word positions. This
confirms that the router learns a difficulty-sensitive signal rather than
uniform layer skipping.

\paragraph{Limitations.}
\begin{itemize}
  \item \emph{Scale.} All experiments used $\approx$5--6M parameters; scaling
    behaviour at 10M--100M is unknown.
  \item \emph{Batch-size dependence.} Sparse-TSA wall-clock speedup required
    batch $\geq 64$; custom Metal kernels would likely eliminate this.
  \item \emph{No attention sparsity.} Only FFN was sparsified; block-sparse
    attention could yield larger savings but would require retraining.
  \item \emph{Preliminary routing analysis.} Figure~\ref{fig:routing_heatmap}
    shows qualitative patterns; quantitative analysis by token type
    (e.g., frequency, entropy) remains future work.
\end{itemize}

% ============================================================
\section{Related Work}

\paragraph{Adaptive Computation.}
\citet{graves2016act} introduced ACT for RNNs; \citet{dehghani2019universal}
extended this approach to weight-shared layers with the Universal Transformer.
\tsa differs in using separate blocks and per-layer soft gates rather than an
accumulated budget. \citet{raposo2024mixture} proposed Mixture-of-Depths, which uses
hard top-$k$ routing; \citet{bae2025mor} introduced Mixture of Recursions, which
applies recursive blocks for a learned number of steps. Both employ discrete routing;
\tsa uses a continuous gate.

\paragraph{Early Exit.}
\citet{elbayad2020depth} proposed the Depth-Adaptive Transformer, which trains
auxiliary classifiers and exits tokens at inference based on output confidence.
Training cost equals the baseline. Our ablation (\S\ref{sec:ablation_ee}) showed
that \tsa achieved better quality at matched efficiency, suggesting that end-to-end
learned routing outperforms post-hoc confidence thresholding.

\paragraph{Other Approaches.}
\citet{chen2025itt} proposed the Inner Thinking Transformer, which augments compute
at hard positions (complementary to \tsa). \citet{fedus2022switch} introduced Switch
Transformers, which route tokens to different expert FFNs, varying width rather than
depth.

% ============================================================
\section{Conclusion}

\tsa reduced token-layer operations by 14--23\% on character-level language modeling
and up to 55\% on synthetic tasks, at $<$0.5\% quality cost across two language
corpora. The router
learns difficulty-proportional allocation from the task-loss gradient alone, producing
meaningful routing even at $\lambda{=}0$. The mechanism adds 1.7\% parameters, proved
robust to $\lambda$ across two orders of magnitude, achieved better quality than early
exit at matched efficiency, and translated to real wall-clock speedup via sparse
inference at batch $\geq 64$. Scaling to 10M+ parameters and per-position routing
analysis are ongoing.

% ============================================================
\bibliography{references}
\bibliographystyle{apalike}

% ============================================================
\appendix
\section{Implementation Details}
\label{app:impl}

\paragraph{Weight Initialisation.}
Token and positional embeddings: $\mathcal{N}(0, 0.02^2)$. Residual projections
(attention output and final feed-forward layer):
$\mathcal{N}(0,\, (0.02 / \sqrt{2L})^2)$, following GPT-3 \citep{brown2020gpt3}.
Router final bias: $-1.0$. Weight tying between token embedding and output head
followed \citet{press2017tying}.

\paragraph{Optimiser.}
AdamW \citep{loshchilov2019adamw} with $\beta = (0.9,\, 0.95)$,
$\lambda_\mathrm{wd} = 0.1$ on all parameters except biases, LayerNorm parameters,
and embeddings (which used $\lambda_\mathrm{wd} = 0$). The MLX implementation applies
weight decay uniformly to all parameters, as MLX does not support per-parameter decay
groups. This affects the enwik8 experiments (Table~\ref{tab:enwik8}) and the $\lambda$
sweep (Table~\ref{tab:lambda_full}).

\paragraph{Causal Masking.}
Standard upper-triangular causal attention mask was used. Routing decisions were
computed from hidden states after the causal attention sublayer and did not depend on
the mask structure.

\paragraph{Character-Level Tokenisation.}
Characters were mapped to integer indices via a sorted vocabulary of unique characters
in the training corpus (65 characters in Tiny-Shakespeare; 6{,}064 unique characters in
enwik8). Token index 0 mapped to the newline character in Shakespeare; the embedding
for this token was initialised normally (no \texttt{padding\_idx} was set) to avoid
zeroing gradients for $\approx$8\% of corpus tokens. enwik8 used the raw byte
distribution with no preprocessing beyond vocabulary construction.

% ============================================================
\section{enwik8 Training Curves}
\label{app:enwik8}

\begin{figure}[h]
  \centering
  \begin{subfigure}[b]{0.48\textwidth}
    \includegraphics[width=\textwidth]{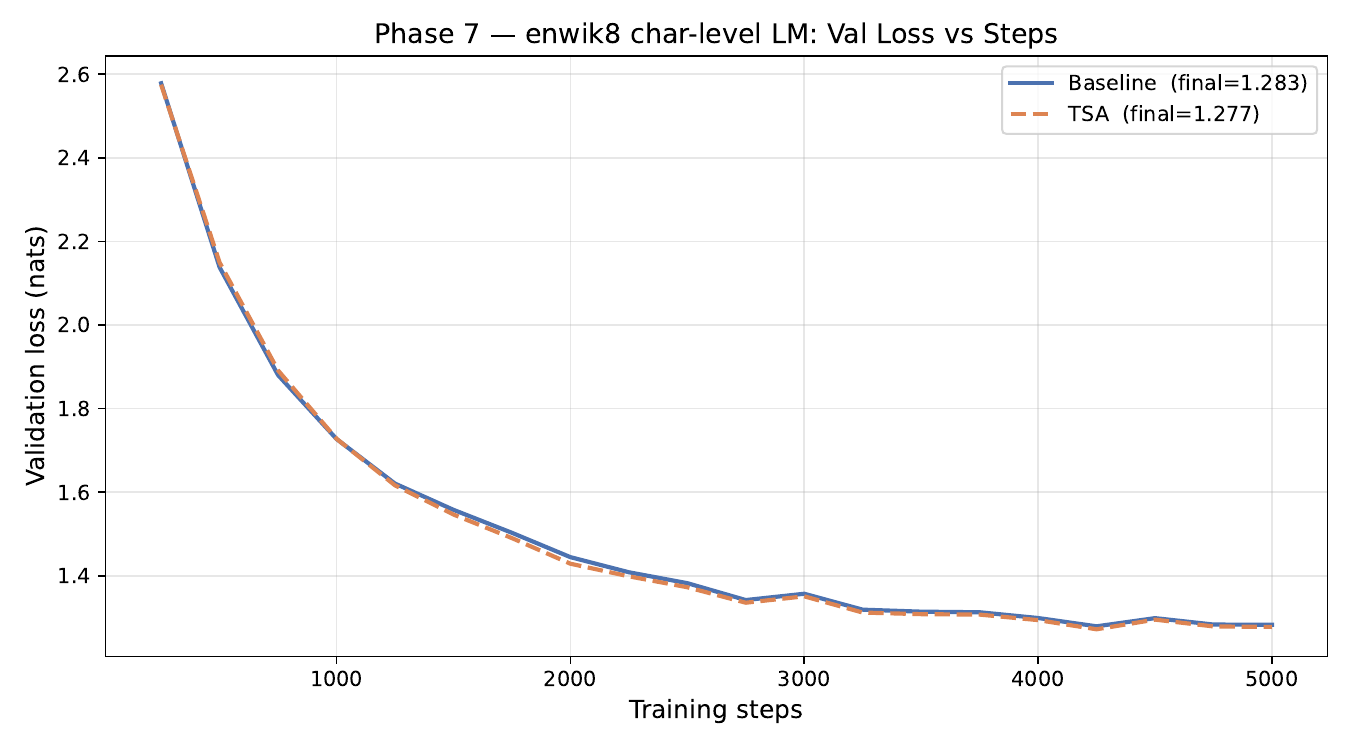}
    \caption{Validation loss vs.\ training step.}
  \end{subfigure}
  \hfill
  \begin{subfigure}[b]{0.48\textwidth}
    \includegraphics[width=\textwidth]{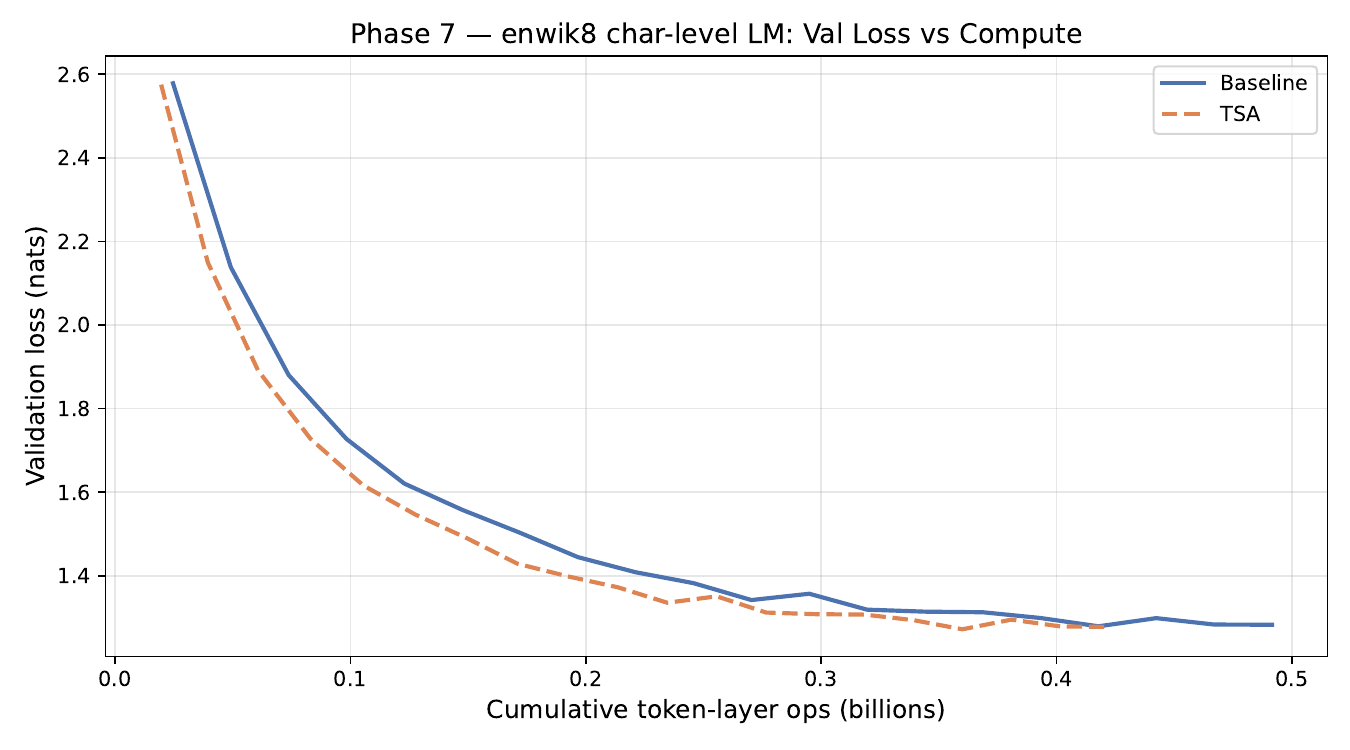}
    \caption{Validation loss vs.\ cumulative \tlops.}
  \end{subfigure}
  \caption{TSA (red) and Baseline (blue) on enwik8. Convergence speed is identical;
  \tsa reaches the same loss for 13.9\% fewer token-layer operations.}
  \label{fig:enwik8_curves}
\end{figure}

% ============================================================
\section{Full $\lambda$ Sweep Results}
\label{app:lambda}

\begin{table}[h]
\centering
\caption{Full $\lambda$ Sweep on Tiny-Shakespeare (5{,}000 Steps, Batch=64, Ctx=128)}
\label{tab:lambda_full}
\small
\begin{tabular}{lccccc}
\toprule
$\lambda$ & Val Loss & BPC & $\alpha$ & TLOps saved & Note \\
\midrule
--- (Baseline) & 1.4422 & 2.0807 & 1.000 & 0\% & baseline \\
0        & \textbf{1.4388} & \textbf{2.0757} & 0.755 & 20.4\% & routes w/o pressure \\
0.001    & 1.4391 & 2.0762 & 0.747 & 21.1\% & default \\
0.005    & 1.4423 & 2.0808 & 0.702 & 24.8\% & \\
0.01     & 1.4412 & 2.0793 & 0.659 & 28.4\% & \\
\textbf{0.05} & \textbf{1.4452} & \textbf{2.0849} & \textbf{0.395} & \textbf{50.4\%} & Pareto-optimal \\
0.1      & 1.4538 & 2.0974 & 0.260 & 61.7\% & \\
0.5      & 1.5271 & 2.2031 & 0.036 & 80.3\% & degraded ($+$5.9\%) \\
\bottomrule
\end{tabular}
\\[3pt]
{\footnotesize All $\lambda$ sweep experiments used MLX on Apple M1~Pro;
within-sweep comparisons are framework-consistent. The $\lambda{=}0.001$ result
differs from Table~\ref{tab:language} (PyTorch, MPS) by 0.63\% in validation loss
due to framework and RNG differences; cross-framework comparisons should use the
sweep-internal baseline row above, not Table~\ref{tab:language}.}
\end{table}

% ============================================================
\section{Full Early Exit Threshold Sweep}
\label{app:early_exit}

\begin{table}[H]
\centering
\caption{Early Exit Threshold Sweep on Tiny-Shakespeare (Post 5{,}000 Steps Training)}
\label{tab:ee_full}
\small
\begin{tabular}{lcccc}
\toprule
Threshold & Val Loss & BPC & $\alpha$ & TLOps saved \\
\midrule
0.99 & 1.4452 & 2.0849 & 0.933 & 5.6\% \\
0.95 & 1.4357 & 2.0713 & 0.864 & 11.3\% \\
0.90 & 1.4471 & 2.0877 & 0.820 & 15.0\% \\
0.85 & 1.4481 & 2.0891 & 0.785 & 17.9\% \\
0.80 & 1.4459 & 2.0861 & 0.756 & 20.3\% \\
\textbf{0.75} & \textbf{1.4586} & \textbf{2.1043} & \textbf{0.728} & \textbf{22.7\%} \\
0.70 & 1.4475 & 2.0883 & 0.700 & 25.0\% \\
0.65 & 1.4439 & 2.0831 & 0.673 & 27.3\% \\
0.60 & 1.4532 & 2.0965 & 0.644 & 29.7\% \\
0.55 & 1.4582 & 2.1037 & 0.615 & 32.0\% \\
0.50 & 1.4727 & 2.1247 & 0.585 & 34.6\% \\
0.40 & 1.4738 & 2.1263 & 0.523 & 39.8\% \\
0.30 & 1.4929 & 2.1538 & 0.462 & 44.9\% \\
\bottomrule
\end{tabular}
\\[3pt]
{\footnotesize Bold row matches the \tsa operating point ($\alpha \approx 0.726$).
Full model (no exit): val loss = 1.4450, identical to Baseline training cost.}
\end{table}

% ============================================================
\section{Full Wall-Clock Data}
\label{app:wallclock}

\begin{table}[h]
\centering
\caption{Active-Fraction Sweep (M1 Pro, MLX, Batch=64, Seq=256)}
\label{tab:wc_full}
\small
\begin{tabular}{lccccc}
\toprule
$\alpha$ & TLOps saved & Baseline (ms) & Sparse-TSA (ms) & Speedup \\
\midrule
0.100 & 75.0\% & 113.2 & 90.8  & $1.246\times$ \\
0.200 & 66.7\% & 113.2 & 93.8  & $1.207\times$ \\
0.300 & 58.3\% & 113.2 & 96.9  & $1.168\times$ \\
0.400 & 50.0\% & 113.2 & 100.3 & $1.129\times$ \\
0.500 & 41.7\% & 113.2 & 103.2 & $1.096\times$ \\
0.600 & 33.3\% & 113.2 & 105.9 & $1.068\times$ \\
0.700 & 25.0\% & 113.2 & 110.3 & $1.026\times$ \\
0.726 & 22.8\% & 113.2 & 110.7 & $\mathbf{1.023\times}$ \\
0.800 & 16.7\% & 113.2 & 112.4 & $1.006\times$ \\
0.833 & 13.9\% & 113.2 & 113.1 & $1.000\times$ \\
0.900 & 8.3\%  & 113.2 & 115.3 & $0.981\times$ \\
1.000 & 0.0\%  & 113.2 & 118.3 & $0.956\times$ \\
\bottomrule
\end{tabular}
\end{table}

\begin{figure}[h]
  \centering
  \includegraphics[width=0.65\textwidth]{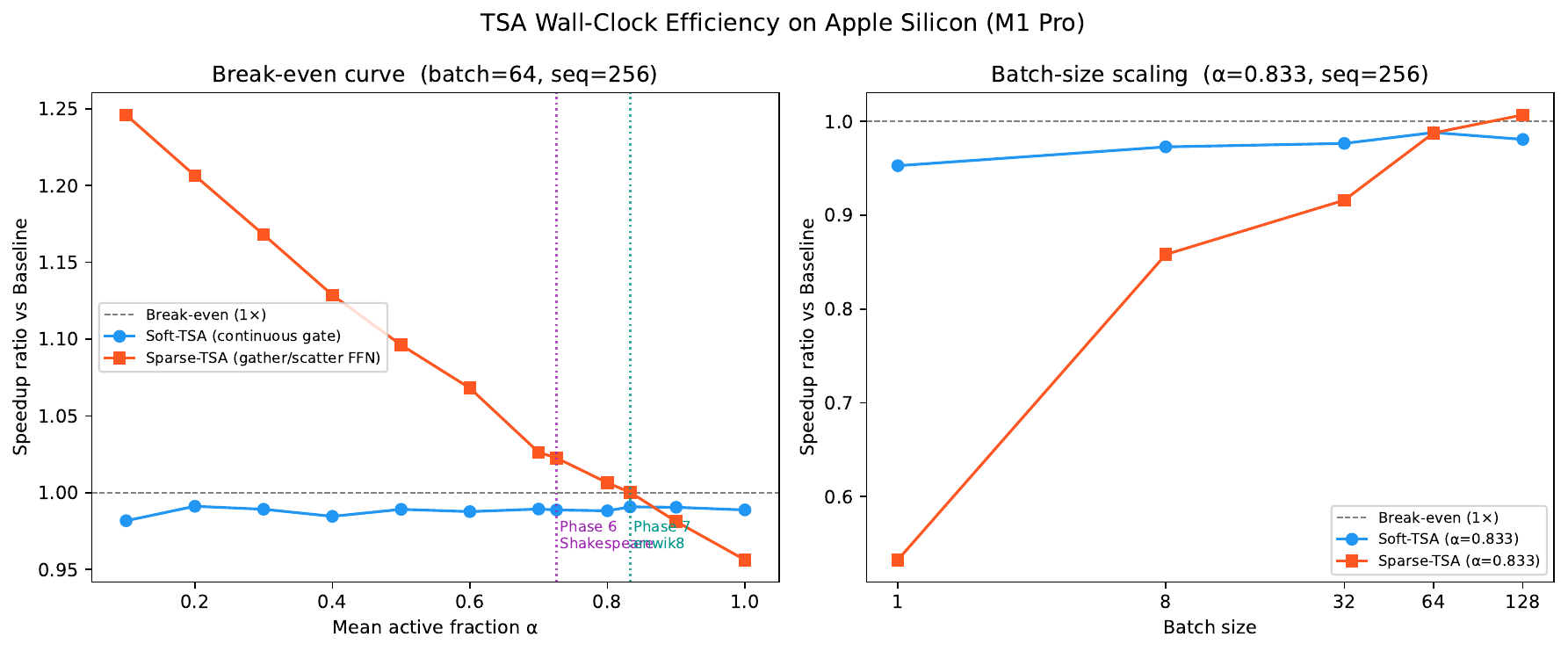}
  \caption{Wall-clock speedup vs.\ active fraction. Sparse-TSA is faster than
  Baseline for $\alpha \leq 0.83$. Vertical dashed lines mark the Shakespeare
  ($\alpha = 0.726$) and enwik8 ($\alpha = 0.833$) operating points.}
  \label{fig:wallclock}
\end{figure}

\begin{table}[h]
\centering
\caption{Batch-Size Scaling at $\alpha = 0.833$ (M1 Pro, MLX, Seq=256)}
\label{tab:batch_scaling}
\small
\begin{tabular}{lcccc}
\toprule
Batch & Baseline (ms) & Sparse-TSA (ms) & Speedup \\
\midrule
1   & 3.3   & 6.2   & $0.533\times$ \\
8   & 16.1  & 18.8  & $0.858\times$ \\
32  & 56.7  & 61.9  & $0.916\times$ \\
64  & 113.2 & 113.1 & $1.000\times$ \\
128 & 225.9 & 224.3 & $1.007\times$ \\
\bottomrule
\end{tabular}
\\[3pt]
{\footnotesize At $\alpha{=}0.833$, sparse-TSA breaks even at batch$=$64 and
achieves speedup at batch$=$128; at batch$=$1, CPU--GPU syncs dominate.}
\end{table}

\end{document}